\documentclass[a4paper, 10 pt, conference]{hsmr} 
\usepackage[utf8]{inputenc}
\IEEEoverridecommandlockouts                       
\usepackage{mathtools}
\usepackage{geometry}
\usepackage[labelsep=space]{caption}
\usepackage{newtxmath,newtxtext}
\usepackage{authblk}
\usepackage{pgfplots}
\usepackage{graphicx}
\usepackage[colorlinks,urlcolor=blue]{hyperref} 
\usepackage{newtxtext,newtxmath}

\pgfplotsset{compat=newest} 
 
\title{\LARGE \bf
Automatic Cannulation of Femoral Vessels in a Porcine Shock Model} 

\author{\LARGE Nico Zevallos$^{1}$,
               Cecilia G. Morales$^{1}$,
               Andrew Orekhov$^{1}$,
               Tejas Rane $^{1}$,\\
               Hernando Gomez$^{2}$,
               Francis X. Guyette$^{2}$,
               Michael R. Pinsky$^{2}$,\\
               John Galeotti$^{1}$,
               Artur Dubrawski$^{1}$,}
\author{\LARGE Howie Choset$^{1}$} 

\affil{\Large\textit{$^{1}$Robotics Institute, Carnegie Mellon University,}\\ \Large\textit{$^{2}$Department of Critical Care Medicine, University of Pittsburgh}\\ \Large\textit{nzevallo@andrew.cmu.edu}}

\begin{document}

\maketitle
\thispagestyle{empty}
\pagestyle{empty}

\newcommand\blfootnote[1]{%
  \begingroup
  \renewcommand\thefootnote{}\footnote{#1}%
  \addtocounter{footnote}{-1}%
  \endgroup
}

\blfootnote{This work was supported by DoD BAA W811XMH-18-SB-AA1 BA180061 and W81XWH-19-C-0101.}

\section*{INTRODUCTION}
Rapid and reliable vascular access is critical in trauma and critical care. Central vascular catheterization enables high-volume resuscitation, hemodynamic monitoring, and advanced interventions like ECMO and REBOA. While peripheral access is common, central access is often necessary but requires specialized ultrasound-guided skills, posing challenges in prehospital settings. The complexity arises from deep target vessels and the precision needed for needle placement. Traditional techniques, like the Seldinger method, demand expertise to avoid complications. Despite its importance, ultrasound-guided central access is underutilized due to limited field expertise. While autonomous needle insertion has been explored for peripheral vessels~\cite{Chen2020Feb}, only semi-autonomous methods~\cite{brattain_ai-enabled_2021} exist for femoral access. This work advances toward full automation, integrating robotic ultrasound for minimally invasive emergency procedures. Our key contribution is the successful femoral vein and artery cannulation in a porcine hemorrhagic shock model.

\section*{MATERIALS AND METHODS}

In our previous work, we developed a system for teleoperated femoral access \cite{zevallos_toward_2021}. This system is an autonomous extension of that work. The system consists of a Universal Robots UR3e arm, which has six degrees of freedom. Attached to the robot arm is a Fukuda 
%FUT-LA-385-12P 
linear ultrasound probe, a 6 axis ATI Nano25 force sensor, an Intel RealSense, 
%D435i
and a needle insertion mechanism composed of a Actuonix L12p linear actuator and a Dynamixel AX-12A servo.

\begin{figure}[t]
    \centering
    \includegraphics[width=.9\columnwidth]{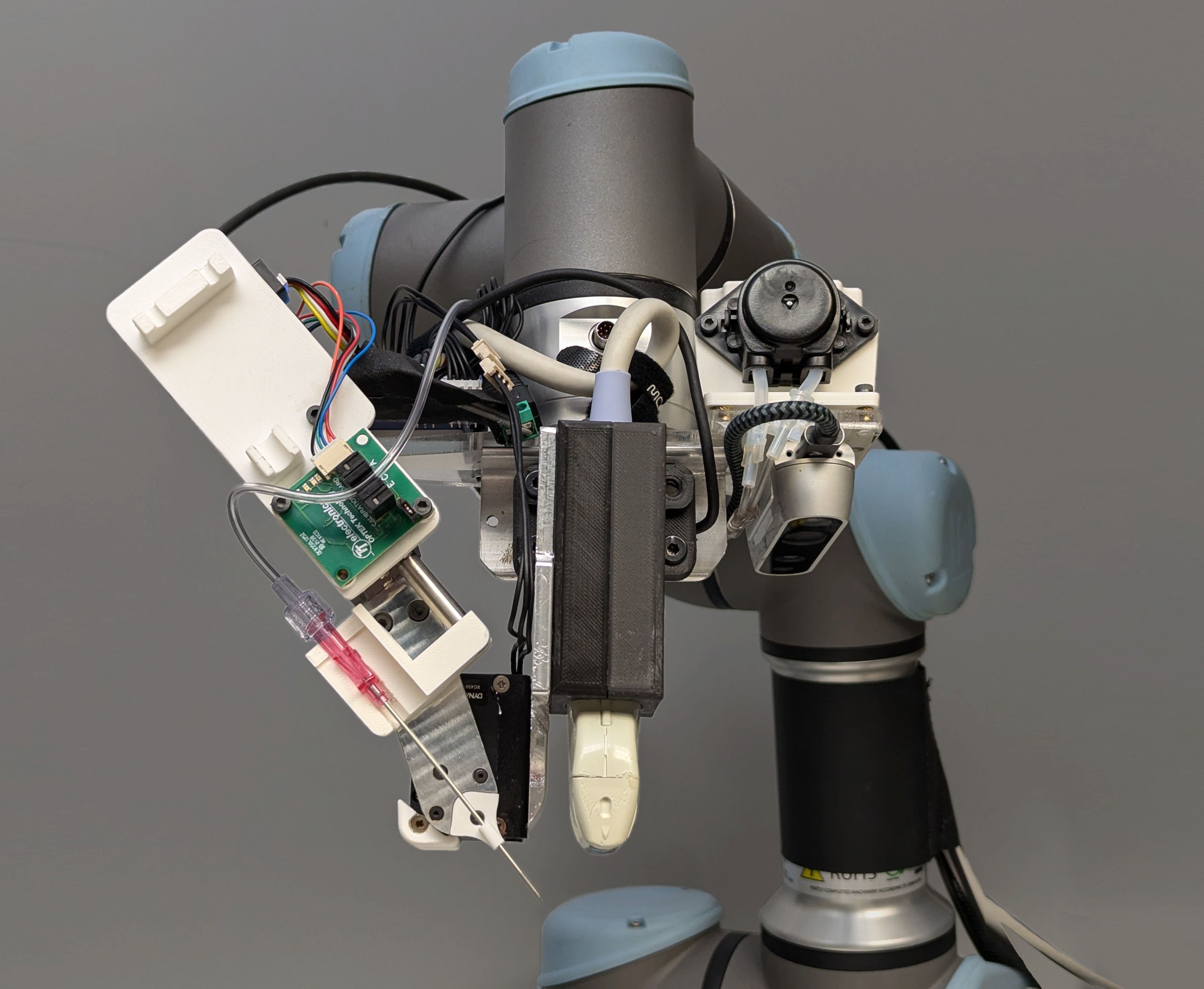}
    \caption{Robotic system}
    \label{fig:pressure_system}
\end{figure}

% To simulate non-compressible bleed in an anesthetized and ventilated porcine model (wt 30 kg), following a midline laparotomy at baseline two 2x2 cm stellate through-and-through right lobe of liver lacerations were performed and the animal observed until mean arterial pressure decreased to $< 40$ mmHg. After which time the abdomen was closed. Once the arterial pressure decreased to $< 10$ mmHg the animal was considered to be in a state of deep hemorrhagic shock and the process proceeded as follows.

To ensure realistic vessel conditions representative of emergency scenarios, our experiments were conducted on live porcine subjects in compliance with the Institutional Animal Care and Use Committee (IACUC) protocol approved by the relevant authority. To model a non-compressible hemorrhage in an anesthetized and ventilated porcine subject weighing 30 kg, a midline laparotomy was performed, followed by the creation of two 2×2 cm stellate through-and-through lacerations in the right lobe of the liver. The animal was monitored until the mean arterial pressure (MAP) dropped below 40 mmHg, at which point the abdomen was closed. Once MAP reached 10mmHg, robotic cannulation of femoral vein and artery were performed in the right groin.

For vascular access, we first capture an RGBD point cloud of the skin surface of the leg using the RealSense camera. The user marks two reference points: one proximally near the hip and the other distally toward the knee. These serve as landmarks for acquiring a detailed scan of the skin surface, following the method described in \cite{Bal_2023}. The resulting dense scan is then cropped using four additional user-defined points to isolate a region just below the inguinal fold, encompassing the femoral artery and vein. At this stage, user input is complete, enabling the automated process to proceed. 

From the cropped point cloud, an ultrasound raster scan path is generated to ensure full coverage of the targeted region, with each pass overlapping by one-third of the probe width. The robot then moves the ultrasound probe along the surface using admittance control~\cite{zevallos_toward_2021}. During scanning, a 2D U-Net segmentation model, trained as described in \cite{Morales_2023}, processes each ultrasound frame in real time to extract vessel contours.

Vessel contours are associated with previously detected contours using the Kuhn–Munkres algorithm, with a cost function based on 2D overlap. These contours are mapped into 3D space using the corresponding robot poses. To reconstruct the vessel paths, splines are fitted to the centroids of the 3D contours, following an approach inspired by \cite{Barratt_2004}. A key adaptation in our method accounts for multiple scans of the same vessel. Instead of parameterizing the vessels as a function of time, we used the projected distance along a linear fit of detected centroids. To simplify representation, a single radius parameter was used instead of complex per-contour curve fitting. Overlapping splines were recursively merged to generate the final vessel reconstruction.

Once the vessels are reconstructed, insertion points are determined along their paths. These possible points are filtered based on reachability and vessel radius. Points were spaced 1 cm apart and ordered from distal to proximal. This arrangement is critical, as arterial trauma triggers vasoconstriction to minimize blood loss, primarily affecting the distal segments to the injury. Prioritizing proximal insertion maximizes the likelihood of successful cannulation even if initial attempts fail. 

Once the robot moves to the first insertion point, the contours of the vessel from automatic segmentation are used to find the target vessel in the ultrasound image. The robot then adjusts the ultrasound probe until the vessel is centered.

Afterwards, a needle insertion mechanism similar to that found in \cite{brattain_ai-enabled_2021} was used to cannulate the vessel, with two key modifications: first, there is an increased probe-to-needle distance, since a steeper insertion angle hindered guide-wire placement, so a greater separation was used to optimize insertion. Second, instead of using a syringe for negative pressure, which introduced air into the line and caused clogging at high pressures, a saline drip maintained fluid presence pre-insertion. A small peristaltic pump controlled negative pressure (-50 mmHg) activating only once the needle was inside the skin. 

Puncture detection was performed using pressure monitoring and blood flashback. Once puncture was detected, the user inserted a preloaded guide-wire into the vessel, and the needle was withdrawn. If insertion failed after dithering the needle, the system moves onto the next best insertion point.

\section*{RESULTS}

We successfully cannulated and placed guide-wires in two live porcine subjects using our system. In the first case, the animal's arterial pressure was 8 mmHg at the time of insertion. Of the six insertion attempts, three were successful: two in the artery and one into the vein. In the second case, the arterial pressure was 2 mmHg at the time of insertion. During the second trial, there was one successful arterial insertion out of two insertion attempts, but the vein was not detected by our segmentation algorithm, so no attempts were made to cannulate it.

\begin{figure}[t]
    \centering
    \includegraphics[width=0.8\columnwidth]{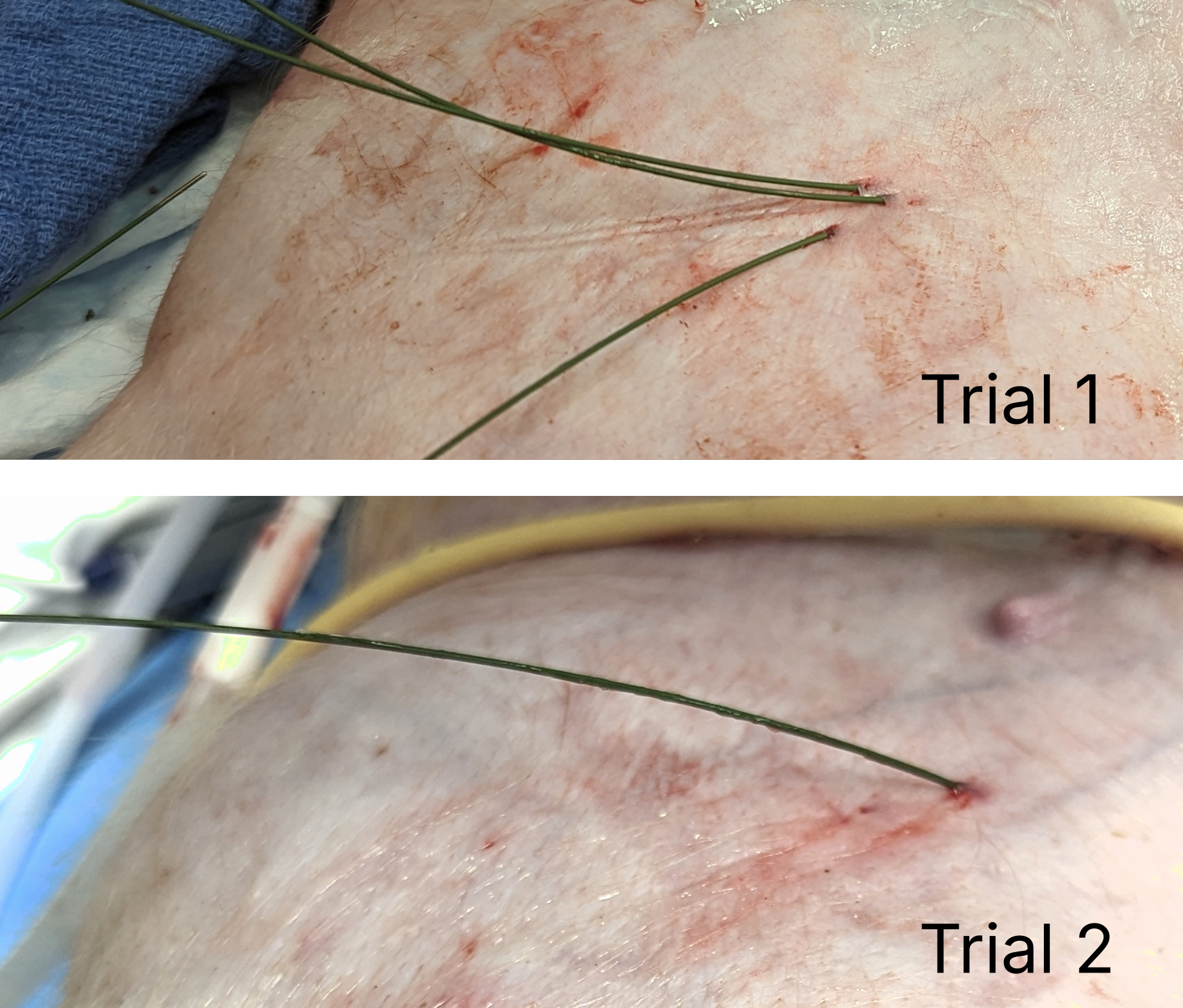}
    \caption{Successful guidewire placements}
    \label{fig:success}
\end{figure}

\section*{DISCUSSION}

While this system was previously extensively evaluated on a phantom, the unique challenges presented by the shock model made insertion significantly more difficult. With such low blood pressure, vessels, particularly veins, collapsed easily during scanning, complicating detection. This was largely due to the segmentation algorithm being trained on animals with normal blood pressure (MAP ~65 mmHg). Additionally, low blood pressure meant more vessel tenting during insertion. Future work should focus on developing more realistic phantoms that better replicate challenges exacerbated by severe hemorrhage.

\nocite{*}
\bibliographystyle{IEEEtran}
\bibliography{HSMR}

\end{document}